# Dealing with Uncertainty in Fuzzy Inductive Reasoning Methodology

**Francisco Mugica**
Centro de Investigación en Ciencia
Aplicada y Tecnología Avanzada
Instituto Politécnico Nacional
Legaria 694, Col. Irrigación
México D.F. 11500, México

**Angela Nebot**
Dept. Llenguatges i
Sistemes Informàtics
Uni. Politècnica de Catalunya
Jordi Girona Salgado, 1-3,
Barcelona 08034, Spain

**Pilar Gómez**
U.P.I.I.C.S.A.
Instituto Politécnico Nacional
The 950, Col. Granjas México
Del. Iztacalco
México D.F. 08400, México

## Abstract

The aim of this research is to develop a strategy of reasoning under uncertainty in the context of the Fuzzy Inductive Reasoning methodology. This methodology allows the prediction of systems behavior by means of two different schemes. The first one corresponds to a *pattern prediction scheme*, based exclusively on pattern rules. The second one corresponds to a purely Sugeno inference system, i.e. *Sugeno prediction scheme*. The Sugeno fuzzy rules are automatically extracted from the pattern rules producing a compact representation of the system modelled. In this paper a mixed pattern/fuzzy rules scheme is studied to deal with uncertainty in such a way that the best of both perspectives is used. The proposed scheme is applied to a real biomedical system, i.e. the central nervous system control of the cardiovascular system.

## 1   INTRODUCTION

The Fuzzy Inductive Reasoning (FIR) methodology emerged from the General Systems Problem Solving (GSPS) developed by Klir [2]. FIR is a data driven methodology based on systems behavior rather than structural knowledge. It is a very useful tool for modelling and simulate those systems from which no previous structural knowledge is available. FIR is composed of four main processes, namely: *fuzzification*, *qualitative model identification*, *fuzzy forecasting*, and *defuzzification*. The *FIR structure* box in figure 1 describes all the processes of FIR methodology.

The fuzzification process converts quantitative data stemming from the system into qualitative data. The model identification process is able to obtain good qualitative relations between the variables that compose the system, building a pattern rule base that guides the fuzzy forecasting process.

The fuzzy forecasting process predicts systems behavior. The FIR inference engine is a specialization of the $k$-nearest neighbor rule, commonly used in the pattern recognition field.

Defuzzification is the inverse process of fuzzification. It makes possible to convert the qualitative predicted output into a quantitative variable that can then be used as input to an external quantitative model. For a deeper insight into FIR methodology the reader is referred to [4, 1].

It has been shown in previous works that FIR methodology is a powerful tool for the identification and prediction of real systems, specially when poor or non structural knowledge is available [3, 5, 6]. However, FIR methodology has an important drawback. The pattern rule base generated by the qualitative model identification process can be very large if there exists a big amount of data available from the system.

In order to solve this drawback it is possible to compact the pattern rule base by extracting Sugeno classical rules from them in an automatic way. The methodology of the automatic construction of fuzzy rules (CARFIR) is the responsible of this process. CARFIR proposes an alternative for the last two processes of FIR methodology (fuzzy forecasting and deffuzification) that consists of a fuzzy inference system (FIS) that allows to compact the pattern rule base in a classical fuzzy rule base and to define an inference scheme that affords the prediction of the future behavior of the system. This is shown in the *FIS structure* box in figure 1. The additional structure does not pretend to substitute the fuzzy prediction and deffuzification processes but to increase the efficiency of FIR methodology.

The extended methodology obtains a fuzzy rule base by means of the *fuzzy rules identification* process that preserves as much information as possible contained in the pattern rule base. Therefore, the former can be considered a generalization of the latter. In other words, the fuzzy rule base is a set of compacted rules that contains the knowledge of the pattern rule base. In this process some precision is lost but robustness is considerably increased.

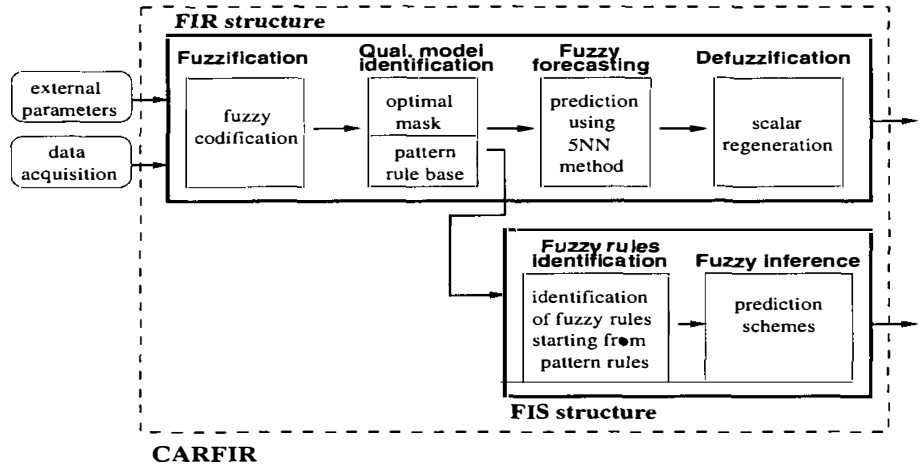

Figure 1: CARFIR structure

The *fuzzy inference* process of CARFIR methodology allows the prediction of systems behavior by means of two different schemes. The first scheme corresponds to the classical forecasting process of FIR methodology, i.e. *pattern prediction scheme*. The second corresponds to purely Sugeno inference system, i.e. *Sugeno prediction scheme*. The pattern prediction scheme is desirable when the computational resources make it possible to deal with the overall pattern rule base or when the extracted fuzzy rules are not accurate enough due to the associated uncertainty. The Sugeno prediction scheme is a good option when the amount of uncertainty associated with the relation between the antecedents and the consequent is small. In this case, the information synthesized in the fuzzy rules allows to obtain very accurate predictions.

In this paper a mixed pattern/fuzzy rules strategy is proposed to deal with uncertainty in such a way that the best of both schemes is used. Areas in the data space with a higher level of uncertainty are identified by means of the so-called error models. The prediction process in these areas makes use of a mixed pattern/fuzzy rules scheme, whereas areas identified with a lower level of uncertainty only use the Sugeno fuzzy rule base.

The performance of the new scheme is studied in the context of a biomedical application, i.e. the human central nervous system control (CNS). The central nervous system is part of the cardiovascular system and controls the hemodynamical system, by generating the regulating signals for blood vessels and heart.

These signals are transmitted through bundles of sympathetic and parasympathetic nerves, producing stimuli in the corresponding organs and other body parts. The CNS control model is composed of five separate controllers: the *heart rate (HR)*, the *peripheric resistance (PR)*, the *myocardiac contractility (MC)*, the *venous tone (VT)*, and the *coronary resistance (CR)*. All of them are single–input/single–output (SISO) models driven by the same input variable, namely the *Carotid Sinus Pressure*.

CARFIR methodology is introduced in section 2. In section 3, the mixed pattern/fuzzy scheme is explained in detail. In section 4 the CNS application is addressed. CARFIR prediction results (pattern, Sugeno and mixed prediction schemes) are presented and discussed from the perspective of the prediction performance and the size of the rule base. Finally, the conclusions of this research are given.

## 2   CARFIR METHODOLOGY

CARFIR methodology is composed of two parts, a FIR structure and a FIS structure (see figure 1). As mentioned earlier CARFIR is an extension of the FIR methodology. Therefore, the first part of CARFIR consists in the generation of the pattern rule base using FIR methodology. To this end, the next steps are required:

- Specification of the external parameters
- Qualitative model identification

The second part of CARFIR methodology consists of the next steps:

- Identification of Sugeno rules starting from pattern rules
- Prediction by means of two different schemes

In the fuzzification process it is necessary to provide the number of classes into which the space is going to be discretized, the landmarks (limits between classes) and the shape of the membership function for each input and output variable. The qualitative model identification process of the CARFIR methodology is responsible of finding spatial and temporal causal relations between variables and, therefore, of obtaining the best qualitative model that represents the system. A FIR model is composed by a so-called *mask* and the *behavior matrix*. The mask represents the structure of the model, whereas the behavior matrix is the associated pattern rule base. These first two steps are explained in detail in [4].

Once the qualitative model identification process is finished, the pattern rule base containing systems behavior is already available. The next step is the generation of fuzzy rules starting from the pattern rules by automatically adjusting the parameters of the fuzzy system. Traditionally, the development of a fuzzy system requires the collaboration of a human expert who is responsible of manually calibrating and tuning all its parameters. It is well known that this is not an easy task and requires a good knowledge of the system.

The CARFIR methodology allows the automatic construction of a fuzzy rule base as a generalization of the previously obtained pattern rule base by means of the *fuzzy rules identification* process (refer to figure 1). The idea behind obtaining fuzzy rules starting from pattern rules is based on the spatial representation of both kind of rules. From a graphic representation perspective, the Sugeno fuzzy rule base can be seen as a completely uniform surface (mesh), due to the fact that a fuzzy inference system generates a unique output value (consequent) for a set of antecedents. The pattern rule base can be viewed graphically as a thick surface. Should the model identified by FIR be a high quality model then the pattern rules form a uniform thin surface in the input-output space. Nevertheless, if the model obtained is not so good, the spatial representation looks as a surface where the thickness of some parts is more significant than that of others. The thickness of the surface means that for a given input pattern (or a set of antecedents) the output variable (or consequent) can take different class values, i.e. the pattern rule base is not deterministic. As mentioned before, the quality of the model is computed by means of an entropy measure that reflects the level of determinism of the state transition matrix associated to the FIR model. A good model is obtained when it has a high level of determinism associated to its rules and all the physical behavior patterns are represented in the model. The spatial representation of such a situation would be a uniform thin surface. The areas with a thick surface will be the ones where the error model will determine a high value and, therefore, it is desirable to use the mixed scheme in these areas.

The tuning process consists in automatically adjusting the mesh built by the fuzzy inference system to the surface obtained from the pattern rules.

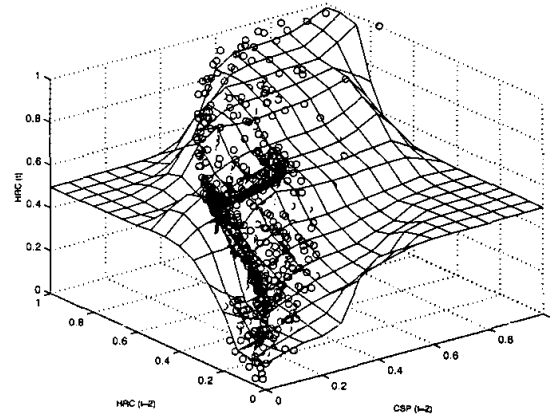

Figure 2: Pattern and fuzzy rules surfaces for the heart rate controller

Figure 2 shows an example of the tuning process. This figure presents a three dimensional view of the graphic representation of the pattern rule base (circles) and the fuzzy rule base (squares) of the heart rate controller of the CNS. The pattern rule base was constructed using the best model inferred by FIR [6]. The output of the Sugeno fuzzy system is obtained by using the product fuzzy operator to the membership values (fires) of the antecedents of each one of the fuzzy rules. The consequents of these rules are the associated weights. Equation 1 is used to perform this process.

$$y = \frac{\sum_{i=1}^{n}(\mu_i \cdot w_i)}{\sum_{i=1}^{n} \mu_i} \qquad (1)$$

In equation 1, $\mu_i$ is the fire of the $i$th fuzzy rule, $w_i$ is the weight of the same rule and $n$ is the total number of Sugeno fuzzy rules. The tuning process consists in adjusting the rules weight, $w_i$, by iterating through the data set using the gradient descent method [7, 8]. The tuning of the $i$th rule weight is obtained by calculating the derivative of the cost function $E$ with respect to $w_i$. The cost function is described in equation 2 (quadratic error addition), where $ND$ is the total number of pattern rules, $y_k$ is the value given by the fuzzy system and $y_k^r$ is the output value of the $k$th pattern rule.

$$E = \frac{1}{2} \sum_{k=1}^{ND} (y_k - y_k^r)^2 \qquad (2)$$

Once the Sugeno pattern rule base has been obtained, it is used to predict the training data set in order to verify its forecasting accuracy.

CARFIR includes two prediction schemes. The first one corresponds to the classical forecasting process of FIR methodology, i.e. *pattern prediction scheme*, that makes use of a specialization of the *k*-nearest algorithm. The second corresponds to a purely Sugeno inference system, i.e. *Sugeno prediction scheme*. In this case the prediction is done by means of the classical Sugeno inference system.

# 3 MIXED PATTERN/FUZZY SCHEME

In the previous section it has been shown how CARFIR synthesizes a Sugeno fuzzy classical model starting from the FIR pattern model previously obtained. In some cases, compacting pattern rules to fuzzy rules could mean a considerable reduction of prediction accuracy. Therefore, it becomes necessary to determine the criteria to be followed about when it is desirable to use the new compacted representation.

CARFIR highly depends on FIR finding a high quality model inherent to which is the uncertainty associated with the relation between the antecedents and the consequent. If the model identified by FIR is a high quality one, the pattern rules will have minimum uncertainty and, therefore, the Sugeno fuzzy rule base will also have a low level of uncertainty. In this case, the prediction accuracy of the fuzzy inference system is similar to the one obtained by the pattern prediction scheme. On the contrary, when the pattern rules have a high level of uncertainty, FIS capacity of capturing the system's knowledge decreases considerably and, in consequence, its prediction accuracy will also decrease. In this case, it is not desirable to use the Sugeno inference system because the fuzzy rule base is unable to be adjusted to the pattern rules surface.

However, the loss of prediction accuracy is not usually present in the whole universe of discourse but only in some specific regions. When this is the case, an interesting option is the design of a mixed inference system that allows to reduce as much as possible the pattern rule base in such a way that only the pattern rules with higher uncertainty are kept.

The use of both approximations requires the identification of an error model that allows to determine what pattern rules to use and when to use them, when to use fuzzy rules and, furthermore, when it is desirable to use a combination of both kinds of rules. In order to select the set of pattern rules that will conform the mixed scheme, it is necessary to make the following considerations:

- FIS scheme loses its prediction accuracy in those regions where the pattern rules surface is thick.

- FIR is able to obtain good predictions in the regions with thickness because it uses the five patterns more similar to the one to be predicted in each moment.

- FIS obtains an acceptable level of accuracy in the areas where the pattern rules surface is thin, replacing a high number of pattern rules and allowing the interpolation in a better way than FIR does.

The error model allows to represent these considerations by determining two bounds, i.e. the minimum error, obtained by the pattern prediction scheme (FIR), and the maximum error, obtained by the fuzzy scheme (FIS) that establishes accuracy reduction in exchange of knowledge compacting.

The main goal of the error model is to find the regions that need to be treated in a mixed form. To this end, it is necessary to compute the quadratic error between the training data set (real) and the predictions obtained by FIR (pattern rules) and FIS (Sugeno fuzzy rules). Three different errors have been considered:

- **G1** Pattern rules (FIR) prediction scheme errors vs. real data

- **G2** Sugeno fuzzy rules (FIS) prediction scheme errors vs. real data

- **G3** Pattern rules (FIR) scheme errors vs. Sugeno fuzzy rules (FIS) scheme errors

These errors can be considered as the distances between the two surfaces that are compared and they can be computed in an accumulative way as average distances for the regions defined by the fuzzy rules antecedents. The cumulative errors obtained for each region allow them to be sorted from higher to lower uncertainty. The idea is to use the mixed scheme for those regions with a higher level of uncertainty, where the pattern rules associated to these regions are kept, whereas the pattern rules associated to the rest of the regions are thrown away and the purely fuzzy scheme is used instead. The percentage of pattern rules kept in the mixed scheme will be determined according to the error reduction impact and will depend on the studied system.

Once the set of pattern rules has been selected, the mixed prediction scheme can take place. The prediction process of the mixed scheme works as follows. The CARFIR methodology generates, on the one hand, a prediction value using the selected fuzzy scheme and, on the other hand, it obtains the prediction value straight from the closest pattern rule. The prediction obtained from the mixed scheme is a weighing of both values. The weighing between these two values is computed with respect to the distance between the antecedents of the system state to be predicted and the antecedents of the closest pattern rule. The similarity between the two sets of antecedents is computed by means of a normalized Euclidean distance measure $d_{real}$, described in equation 3,

$$d_{real} = \sqrt{\sum_{i=1}^{N}(x_i - y_i)^2} \qquad (3)$$

where $x_i$ and $y_i$ are the ith antecedents of the system state and the closest pattern rule, respectively. Although the system variables are already normalized between [0 1], it is necessary to re-normalize the input space with respect to the maximum Euclidean distance that is computed as the square root of the number of antecedents (equation 4) that form the prediction system.

$$d_{max} = \sqrt{n_{antec}} \qquad (4)$$

Therefore, the normalized Euclidean distance, $d_{norm}$, between the antecedents of the point to be predicted and the antecedents of the closest pattern rule can be computed with equation 5,

$$d_{norm} = \frac{d_{real}}{d_{max}} \qquad (5)$$

and it will be the input argument for the $f_{mix}$ function that allows to establish the percent weight of the pattern rules with respect to the fuzzy rules scheme, as shown in equation 6.

$$f_{mix} = \frac{1}{1 - e^{-d_{norm}}} \qquad (6)$$

The $f_{mix}$ function can be adjusted between a minimum and a maximum value and can be redefined in pieces in such a way that for values smaller or equal to 0.01, $f_{mix}$ takes the value of 1 and for values bigger or equal than 0.25, $f_{mix}$ takes the value of 0. The graphical representation of the $f_{mix}$ function is presented in figure 3.

Using this function the mixed prediction scheme can be computed as described in equation 7.

$$y_{mix} = y_{pattern} \cdot f_{mix} + y_{Sugeno} \cdot (1 - f_{mix}) \qquad (7)$$

where $y_{pattern}$ is the output obtained with the pattern prediction scheme and $y_{Sugeno}$ is the output obtained with the Sugeno fuzzy prediction scheme.

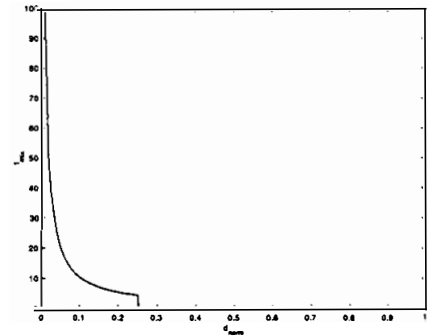

Figure 3: Function used to integrate the mixed pattern/fuzzy scheme

## 4 CNS CONTROLLER MODELS

In this work the five CNS controller models of the cardiovascular system, namely, *heart rate*, *peripheric resistance*, *myocardiac contractility*, *venous tone* and *coronary resistance*, are inferred for a specific patient by means of CARFIR methodology.

As has been mentioned earlier, all the controllers are SISO models driven by the same input variable, the *carotid sinus pressure*. The input and output signals of the CNS controllers were recorded with a sampling rate of of 0.12 seconds from simulations of the purely differential equation model [6]. The model had been tuned to represent a specific patient suffering a coronary arterial obstruction, by making the four different physiological variables (right auricular pressure, aortic pressure, coronary blood flow, and heart rate) of the simulation model agree with the measurement data taken from the real patient. The five models were validated by using them to forecast six data sets not employed in the training process. Each one of these six test data sets, with a size of about 600 data points each, contains signals representing specific morphologies, allowing the validation of the model for different system behaviors.

In the forecasting process, the normalized mean square error (in percentage) between the predicted output, $\hat{y}(t)$, and the system output, $y(t)$, is used to determine the validity of each of the control models. The error is given in equation 8.

$$MSE = \frac{E[(y(t) - \hat{y}(t))^2]}{y_{var}} \cdot 100\% \qquad (8)$$

where $y_{var}$ denotes the variance of $y(t)$.

The quantitative data obtained from the system is converted into qualitative data by means of the fuzzification process of CARFIR methodology (FIR structure). Several experiments were done with different partitions of the data for the five controllers. Both the input and output variables were discretized into 3, 5, 7 and 9 classes using the equal frequency partition (EFP) method. The identification of the models was carried out using 7277 samples.

Applying the best models found to the qualitative data, a pattern rule base with 7275 rules was obtained for each one of the five controllers. For a deeper insight on the identification of CNS models and the generation of the pattern rule bases refer to [6]. Once the pattern rules are available the fuzzy rules identification procedure can take place. From the experiments performed with different number of classes, it was concluded that the best matching between pattern and fuzzy rules was obtained when the input and output variables were discretized into 9 classes. Therefore, each controller has associated a Sugeno rule base of 81 rules. The reduction of the number of rules is significant (from 7275 to 81). Figure 2 shows a three dimensional view of the graphic representation of the pattern rule base (circles) and the fuzzy rule base (squares) of the heart rate controller of the CNS, obtained after the tuning process. The tuning process has been performed during 50 epochs for all the five controllers.

As can be seen from figure 2, the mesh that represents the fuzzy rules has been adapted quite accurately to the pattern rules surface. This is due to the fact that the thickness of the pattern rules surface is considerably small making the approximation by means of a fuzzy rules surface (mesh) possible.

Once the Sugeno rule base is available for each controller, the Sugeno prediction scheme is performed for each of the 6 test data sets. The $MSE$ errors of the five controller models for each of the test data sets are presented in table 1. The columns of table 1 contain the mean square errors obtained when the 6 test data sets (DS) were predicted using each of the five CNS controllers. The last row of the table shows the average prediction error of the 6 tests for each controller.

Table 1: MSE prediction errors of the CNS controller models using the Sugeno prediction scheme of CARFIR methodology

|  | HR | PR | MC | VT | CR |
|---|---|---|---|---|---|
| DS 1 | 4.92% | 10.91% | 5.54% | 5.56% | 2.05% |
| DS 2 | 4.21% | 9.71% | 4.34% | 4.36% | 2.62% |
| DS 3 | 4.54% | 7.54% | 3.63% | 3.64% | 2.33% |
| DS 4 | 3.24% | 6.23% | 1.47% | 1.46% | 3.87% |
| DS 5 | 4.75% | 9.40% | 8.66% | 8.65% | 2.34% |
| DS 6 | 4.25% | 14.68% | 5.42% | 5.43% | 4.28% |
| Ave. | 4.32% | 9.74% | 4.84% | 4.85% | 2.91% |

From table 1 it can be seen that the coronary resistance (CR) model captures in a reliably way the behavior of this controller, achieving an average error of 2.91%. The largest average error is 9.74% obtained with the peripheric resistance controller (PR) model. Therefore, the PR model is the one that captures less accurately the behavior of the controller.

The results obtained when using the pattern prediction scheme of CARFIR methodology are shown in table 2. As expected, the predictions done using the pattern rule base are much more accurate than the ones obtained by the Sugeno fuzzy rule base.

Table 2: MSE prediction errors of the CNS controller models using the pattern prediction scheme of CARFIR methodology

|  | HR | PR | MC | VT | CR |
|---|---|---|---|---|---|
| DS 1 | 1.49% | 1.29% | 1.06% | 1.06% | 1.03 % |
| DS 2 | 1.45% | 1.34% | 1.11% | 1.11% | 0.87% |
| DS 3 | 1.63% | 1.19% | 1.30% | 1.31% | 0.76 % |
| DS 4 | 1.07% | 0.85% | 0.88% | 0.88% | 0.21% |
| DS 5 | 1.55% | 3.45% | 3.30% | 3.29% | 0.62 % |
| DS 6 | 2.15% | 1.65% | 1.33% | 1.32% | 0.50 % |
| Ave. | 1.56% | 1.63% | 1.49 % | 1.49% | 0.66% |

As can be seen from table 2 the results obtained are very good, with MSE errors lower than 1.7% for all the controllers. The average error obtained for all the controllers is 1.36% much lower than the 5.33% obtained with the Sugeno prediction scheme. Clearly, the prediction capability of the fuzzy rule base is inferior than that of the pattern rule base. It is important to notice that the size of the rule base has been extremely reduced, i.e. from 7275 pattern rules to 81 fuzzy rules. This is a relevant aspect that should be taken into account in the context of the CARFIR methodology.

Table 3: MSE prediction errors of the CNS controller models using NARMAX, TDNN and RNN methodologies

|  | HR | PR | MC | VT | CR |
|---|---|---|---|---|---|
| NARMAX | 9.3% | 18.5% | 22.0% | 22.0% | 25.5% |
| TDNN | 15.3% | 33.7% | 34.0% | 34.0% | 55.6% |
| RNN | 18.3% | 31.1% | 35.1% | 34.7% | 57.1% |

Table 3 contains the predictions achieved when NARMAX, time delay neural networks and recurrent neural networks are used for the same problem. The columns of the table specify the average prediction error of the 6 test sets for each controller. All methodologies used the same training and test data sets previously described. The errors obtained for all the controllers using NARMAX models are larger than the ones obtained by the fuzzy prediction scheme of CARFIR methodology (see table 1). The average prediction error for all the controllers is 19.46% versus the 5.33% accomplished by CARFIR (Sugeno fuzzy scheme). However, NARMAX models are more precise than time delay and recurrent neural networks. The average prediction error computed by TDNNs for the five controllers is 34.57% and 35.3% for RNNs, both larger than the 19.46% obtained by NARMAX models. In [6] the results obtained by NARMAX models were considered acceptable from the medical point of view. In extension, also pattern and fuzzy models of CARFIR methodology should be acceptable, due to their higher prediction performance.

At this point it is interesting to establish a mixed scheme that preserves a small subset of pattern rules and the overall Sugeno fuzzy rule base in such a way that a compromise is found between prediction accuracy and rules compacting. The mixed scheme is applied to the central nervous system control by using the three error models described in the previous section. The incidence in the prediction accuracy of preserving 10%, 20%, 30%, 40% and 60% of the pattern rule base is analyzed. The mean errors obtained for the six test data sets for the coronary resistance controller when the mixed scheme is used are shown in table 4. The columns specify the percentage of pattern rules preserved and the rows determine the error model used. It is interesting to remember here that the error obtained using the pattern prediction scheme was of 1.63% and of 9.74% for the Sugeno prediction scheme. Therefore, a considerable reduction of prediction accuracy (from 1.63% to 9.74%) is derived from the compacting strategy (from 7275 pattern rules to 81 fuzzy rules).

Table 4: MSE mean prediction errors of the PR controller using the mixed scheme

| Mixed | 10% | 20% | 30% | 40% | 60% |
|---|---|---|---|---|---|
| G1 | 8.22% | 6.89% | 5.68% | 4.84% | 3.94% |
| G2 | 5.13% | 3.26% | 2.89% | 2.16% | 1.81% |
| G3 | 5.13% | 3.60% | 2.89% | 2.22% | 1.86% |
| FIR | 1.63% | | FIS% | 9.74% | |

In table 4 it can be seen that the mixed scheme allows to recover prediction accuracy as the number of preserved pattern rules is increased. Looking closer to the G2 error model, it is interesting to notice that with 10% of preserved pattern rules the accuracy loss is reduced to 57%. With 20% of preserved pattern rules the prediction accuracy loss is reduced to 80%.

With respect to the three error models studied in this paper, G2 is clearly the one that obtains better results. G3 results are closer to the ones obtained by G2. Finally, G1 (FIR vs. real) is the one that performs worse, due to the fact that FIS scheme is not considered in this model and, therefore, the accuracy reduction is not measured at all in this case.

Figure 4 graphically shows the prediction recovery as the percentage of pattern rules increases for the PR controller. The dashed lines correspond to the prediction performed by the pattern, mixed (preserving 60% and 20% of the pattern rule base) and Sugeno schemes. The solid lines correspond to the true measured output.

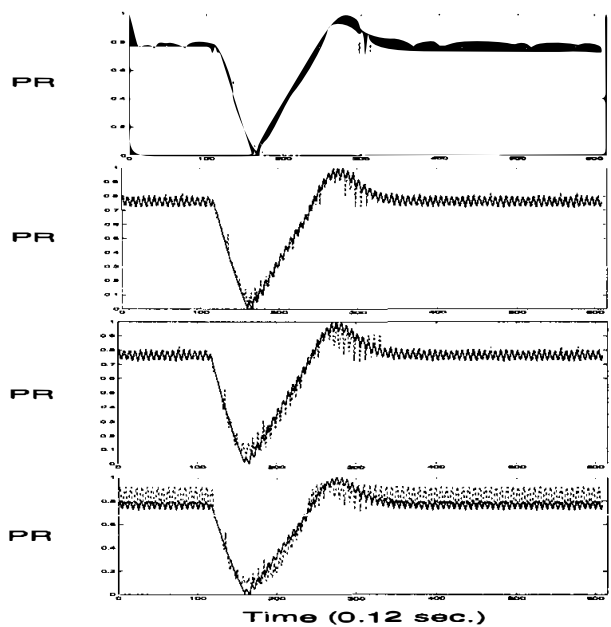

Figure 4: PR controller predictions (test data set #6) using from top to bottom Pattern, Mixed 60%, Mixed 20% and Sugeno schemes

Table 5: MSE mean prediction errors of the MC controller using the mixed scheme

| Mixed | 10% | 20% | 30% | 40% | 60% |
|---|---|---|---|---|---|
| G1 | 4.31% | 3.97% | 3.52% | 3.29% | 2.93% |
| G2 | 3.10% | 2.33% | 2.14% | 1.94% | 1.89% |
| G3 | 3.09% | 2.38% | 2.20% | 2.00% | 1.94% |
| FIR | 1.50% | | FIS% | 4.84% | |

From figure 4 it can be seen that when less pattern rules are used, the accuracy on the details (basically on high frequencies) of the signal diminishes. Notice that the prediction obtained when only Sugeno fuzzy rules are used has also some problems forecasting the upper and lower picks of the signal (low frequency). However, the use of the mixed scheme preserving only a 20% of the pattern rule base can solve this limitation.

The results for the MC, VT, CR and HR controllers are presented in tables 5, 6, 7 and 8, respectively. Previous comments referred to the PR controller are fully applicable to the rest of the CNS controllers.

Table 6: MSE mean prediction errors of the VT controller using the mixed scheme

| Mixed | 10% | 20% | 30% | 40% | 60% |
|---|---|---|---|---|---|
| G1 | 4.09% | 3.50% | 3.14% | 3.01% | 2.71% |
| G2 | 3.10% | 2.34% | 2.14% | 1.94% | 1.89% |
| G3 | 1.42% | 1.18% | 1.08% | 0.98% | 0.95% |
| FIR | 1.49% | | FIS% | 4.85% | |

Table 7: MSE mean prediction errors of the CR controller using the mixed scheme

| Mixed | 10% | 20% | 30% | 40% | 60% |
|-------|-----|-----|-----|-----|-----|
| G1    | 2.95% | 2.98% | 2.89% | 2.79% | 2.23% |
| G2    | 1.51% | 0.90% | 0.67% | 0.63% | 0.51% |
| G3    | 2.91% | 2.07% | 1.32% | 0.82% | 0.57% |
| FIR   | 0.67% |       | FIS% | 2.92% |     |

From tables 4, 5, 6, 7 and 8 it can be seen that there is an asymptotic recovery accuracy behavior as the percentage of pattern rules used in the mixed scheme is increased. For the application at hand, it can be concluded that a 30% of accuracy recovery (or a 70% of pattern rule base reduction) is an optimum value in terms of accuracy/computational cost.

Table 8: MSE mean prediction errors of the HR controller using the mixed scheme

| Mixed | 10% | 20% | 30% | 40% | 60% |
|-------|-----|-----|-----|-----|-----|
| G1    | 3.88% | 3.55% | 3.10% | 2.56% | 1.94% |
| G2    | 3.00% | 2.47% | 2.15% | 1.87% | 1.67% |
| G3    | 3.27% | 2.59% | 2.33% | 2.25% | 1.90% |
| FIR   | 1.56% |       | FIS% | 4.32% |     |

It is important to notice that it is very difficult to define a general decision procedure which allows to decide the percentage of pattern rules to keep in the mixed scheme due to the fact that this decision highly depends on both the application at hand and the users interests. It is not always useful or convenient to use a fully Sugeno prediction scheme or a mixed pattern/fuzzy rules scheme. On the one hand, if there is a high level of uncertainty in all the regions of the model, the use of a Sugeno prediction scheme is definitely not a good idea while the use of a pattern prediction scheme will be absolutely convenient. On the other hand, if all pattern rules are represented by the Sugeno fuzzy rules without losing precision, the Sugeno prediction scheme will be clearly the best choice. In between both extremes, the mixed scheme becomes necessary depending on the purpose of the model. From this perspective, it is important to decide the relation precision vs. speed for the application at hand. For instance, in some applications it is really crucial to obtain a very accurate prediction of the future behavior whereas the time needed to reach it is not the important point. In this case a classical pattern prediction scheme should be the right decision. Contrarily, when the main goal is to obtain a quick prediction and no high level accuracy is needed, then a fully Sugeno prediction scheme is certainly the best alternative. When neither of both requirements is crucial, the mixed scheme allows the user to establish the best compromise taking into account the purpose of the model.

## 5   CONCLUSIONS

In this paper a mixed prediction scheme has been designed that allows to obtain a better compromise between prediction performance and size of the rule base. The mixed scheme is a combination of the Sugeno rules and a reduced set of pattern rules. The advantage of the pattern rules is that they are more accurate than the fuzzy rules in those areas where a large degree of uncertainty exists. In order to take advantage of this fact, the mixed scheme keeps a percentage of pattern rules that allows the prediction of those system states with a high degree of uncertainty. The new scheme has been used to model and predict the human Central Nervous System, showing the potentiality of CARFIR methodology. The next step is to use CARFIR in applications with a higher volume of data, i.e. a bigger number of pattern rules, to see the real potentiality of the methodology. A theoretical study of the size rule bases reduction is expected to be performed in the near future.

### Acknowledgements

This research was supported by Spanish Consejo Interministerial de Ciencia y Tecnología (CICYT), under project DPI2002-03225.